\documentclass[journal]{IEEEtran}

\usepackage{amssymb}
\usepackage{graphicx}
\usepackage{graphicx}
\usepackage{amsfonts}
\usepackage{amsmath}
\usepackage{subcaption}
\usepackage[table,dvipsnames]{xcolor}
\newcolumntype{L}{@{}>{\kern\tabcolsep}l<{\kern\tabcolsep}}
\usepackage{booktabs}
\usepackage{multirow}
\usepackage{enumitem}
\usepackage[misc]{ifsym}
\usepackage[labelfont=bf]{caption}
\usepackage{algorithm,algorithmic}
\usepackage[table]{xcolor}
\newcolumntype{L}{@{}>{\kern\tabcolsep}l<{\kern\tabcolsep}}
\usepackage[pages=some,placement=top]{background}

\usepackage{float}  % allows for indexgeneration

\usepackage{xpatch,letltxmacro}% http://ctan.org/pkg/etoolbox

\DeclareMathOperator*{\argminB}{argmin}

\newcolumntype{P}[1]{>{\centering\arraybackslash}p{#1}}

\usepackage{url}
\urldef{\mailsa}\path|{******, *******, *******, *******,|
\urldef{\mailsb}\path|********, *******, *******, *******,|
\urldef{\mailsc}\path|********, *********, *****}*******|

\DeclareMathOperator*{\minimum}{min}

\makeatletter

\newcommand{\ALC@it@nostep}{\item[]}% No-step algorithmic item
\LetLtxMacro\oldalgorithmic\algorithmic
\LetLtxMacro\endoldalgorithmic\endalgorithmic
\renewenvironment{algorithmic}[1][0]{%
  \oldalgorithmic[#1]%
  \xpatchcmd{\STATE}{\ALC@it}{\ALC@it@nostep}{}{}%
  \xpatchcmd{\FOR}{\ALC@it}{\ALC@it@nostep \textbf{\textit{Step 1 -- }}}{}{}%
  \xpatchcmd{\RETURN}{\ALC@it}{\ALC@it@nostep}{}{}%
  \xpatchcmd{\ENDFOR}{\ALC@it}{\ALC@it@nostep}{}{}%

}{\endoldalgorithmic}
\makeatother

\backgroundsetup{contents=Accepted for publication in IEEE Transactions on Medical Imaging 2019,color=black!100,scale=1.25,opacity=0.7,position={6.75,1.75}}
\BgThispage

\begin{document}

%\backgroundsetup{contents=Under Review for IEEE Transactions on Medical Imaging 2018,color=black!100,scale=1.25,opacity=0.7,position={6.75,1.75}}
%\BgThispage
\title{Lung and Pancreatic Tumor Characterization in the Deep Learning Era: Novel Supervised and Unsupervised Learning Approaches }

\author{Sarfaraz Hussein, Pujan Kandel, Candice W. Bolan, Michael B. Wallace,~and~Ulas~Bagci$^*$, \IEEEmembership{Senior Member,~IEEE.}
  \thanks{$^*$ indicates corresponding author (ulasbagci@gmail.com).}%
 \thanks{S. Hussein is with Center for Advanced Machine Learning (CAML) at Symantec Corporation. U. Bagci is with Center for Research in Computer Vision (CRCV) at University of Central Florida (UCF), Orlando, FL; P. Kandel, C. Bolan, and M. Wallace are with Mayo Clinic, Jacksonville, FL}}

\maketitle

\begin{abstract}
%Cancer is among the leading causes of death worldwide. 
Risk stratification (characterization) of tumors from radiology images can be more accurate and faster with computer-aided diagnosis (CAD) tools. % are often needed for fast and accurate detection, characterization, and risk assessment of different tumors from radiology images. 
Tumor characterization through such tools can also enable non-invasive cancer staging, prognosis, and foster personalized treatment planning as a part of precision medicine. 
In this study, we propose both supervised and unsupervised machine learning strategies to improve tumor characterization. Our \textit{first approach} is based on supervised learning for which we demonstrate significant gains with deep learning algorithms, particularly by utilizing a 3D Convolutional Neural Network and Transfer Learning. 
Motivated by the radiologists' interpretations of the scans, we then show how to incorporate task dependent feature representations into a CAD system via a graph-regularized sparse Multi-Task Learning (MTL) framework. 

In the \textit{second approach}, we explore an unsupervised learning algorithm to address the limited availability of labeled training data, a common problem in medical imaging applications. Inspired by learning from label proportion (LLP) approaches in computer vision, we propose to use proportion-SVM for characterizing tumors. We also seek the answer to the fundamental question about the goodness of ``deep features" for unsupervised tumor classification. 
We evaluate our proposed supervised and unsupervised learning algorithms on two different tumor diagnosis challenges: lung and pancreas with 1018 CT and 171 MRI scans, respectively, and obtain the state-of-the-art sensitivity and specificity results in both problems. 

\end{abstract}


\IEEEpeerreviewmaketitle
%\linenumbers

\section{Introduction}
%\subsection{Central Obesity}
Approximately 40\% of people will be diagnosed with cancer at some point during their lifetime with an overall mortality of 171.2 per 100,000 people per year (based on deaths between 2008-2012)~\cite{americancancer}. Lung and pancreatic cancers are two of the most common cancers. While lung cancer is the largest cause of cancer-related deaths in the world, pancreatic cancer has the poorest prognosis with a 5-year survival rate of only 7\% in the United States~\cite{americancancer}. %Lung cancer accounts for the highest number of mortalities i.e. 1.59 million~\cite{americancancer} out of 8.2 million deaths due to cancer worldwide. 
With regards to pancreatic cancer, specifically in this work, we focus on the challenging problem of automatic diagnosis of Intraductal Papillary Mucinous Neoplasms (IPMN). IPMN is a pre-malignant condition and if left untreated, it can progress to invasive cancer. IPMN is mucin-producing neoplasm that can be found in the main pancreatic duct and its branches. They are radiographically identifiable precursors to pancreatic cancer~\cite{sadot2015tumor}. %~\cite{shi2012intraductal,sadot2015tumor}. 
Detection and characterization of these lung and pancreatic tumors can aid in early diagnosis; hence, increased survival chance through appropriate treatment/surgery plans. 

\begin{figure}[t]
\centering
%\hspace{-0. cm}
\includegraphics[width=90 mm]{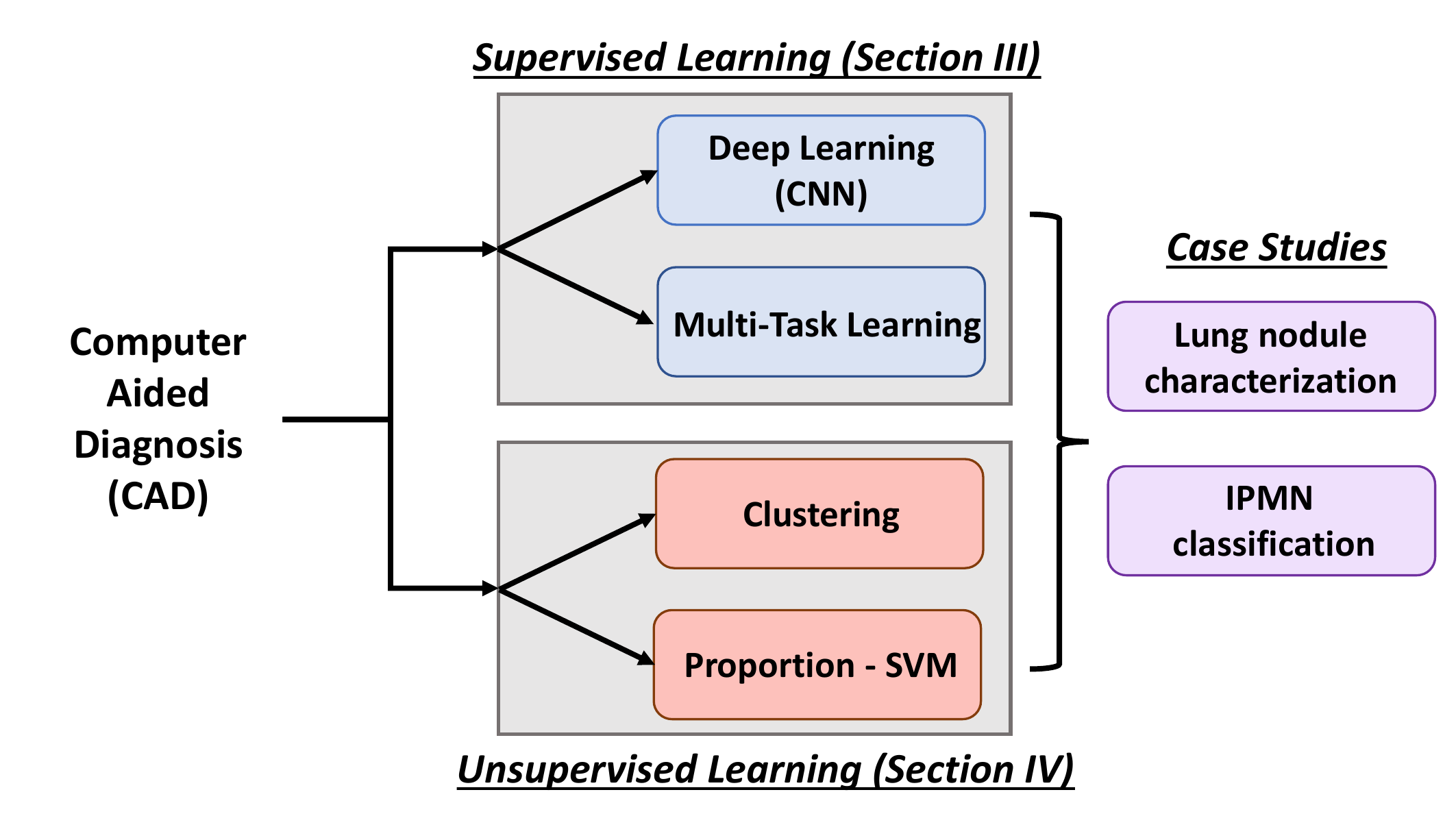}
\caption{A block diagram to represent different schemes, methods and experimental case studies presented in this paper. We develop both  supervised and unsupervised learning algorithms to characterize tumors. For the supervised learning scheme, we propose a new  3D CNN architecture based on a Graph Regularized Sparse Multi-task learning and perform evaluations for lung nodule characterization from CT scans. For unsupervised learning scheme, we propose a new clustering algorithm, $\propto$SVM, and test it for the categorization of lung nodules from CT scans and pancreatic cysts (IPMN) from MRI scans.}
\label{fig:blockdiag}
%\vspace{-0.4 cm}
\end{figure}

Conventionally, the CAD systems are designed to assist radiologists in making accurate and fast decisions by reducing the number of false positives and false negatives. For diagnostic decision making, a higher emphasis is laid on increased sensitivity: a false-flag is more tolerable than a tumor being missed or incorrectly classified as benign. In this regard, a computerized analysis of imaging features becomes a key instrument for radiologists to improve their diagnostic decisions. In the literature, automated detection and diagnosis methods had been developed for tumors in different organs such as breast, colon, brain, lung, liver, prostate, and others. As typical in such studies, a CAD includes preprocessing and feature engineering steps (including feature extraction and selection) followed by a classification step~\cite{el20113d,han2015texture,way2006computer,lee2010computer}. However, with the success of deep learning, a transition from feature engineering to feature learning has been observed in medical image analysis literature. Those systems comprise Convolutional Neural Networks (CNN) as feature extractor followed by a conventional classifier such as Random Forest (RF)~\cite{kumar2015lung,buty2016characterization}. In scenarios where a large number of labeled training examples are available, however, end-to-end trainable deep learning approaches can be employed~\cite{saouli2018fully}.

This paper includes two main approaches for tumor characterization from radiology scans: supervised and unsupervised learning algorithms. In the first part, we focus on novel supervised algorithms, which is a significant extension to our IPMI 2017 study~\cite{hussein2017risk}. Specifically, we first present a novel supervised learning strategy to perform risk-stratification of lung nodules from low-dose CT scans. For this strategy, we perform a 3D CNN based discriminative feature extraction from radiology scans. We contend that 3D networks are important for the characterization of lung nodules in CT images which are inherently 3-dimensional. The use of conventional 2D CNN methods, whereas, leads to the loss of vital volumetric information which can be crucial for precise risk assessment of lung nodules.
In the absence of a large number of labeled training examples, we utilize a pre-trained 3D CNN architecture and fine-tune the network with lung nodules dataset. Also, inspired by the significance of lung nodule attributes for clinical determination of malignancy~\cite{furuya1999new}, we utilize the information about six high-level nodule attributes such as calcification, spiculation, sphericity, lobulation, margin, and texture (Figure~\ref{fig:supworkflow}-A) to improve automatic benign-malignant classification. Then, we integrate these high-level features into a novel graph regularized multi-task learning (MTL) framework to yield the final malignancy output.
We analyze the impact of the aforementioned lung nodule attributes in-depth for malignancy determination and find these attributes to be complementary when obtaining the malignancy scores. From a technical perspective, we also exploit different regularizers and multi-task learning approaches such as trace-norm and graph regularized MTL for regression.

%We also explore the potential of unsupervised learning for lung nodule and IPMN classification. 
In the second part of the paper, inspired by the successful application of unsupervised learning methods in other domains, we explore the potential of unsupervised learning strategies in lung nodule and IPMN classification. First, we extract discriminative information from a large amount of unlabeled imaging data. We analyze both hand-crafted and deep learning features and assess how good those features are when applied to tumor characterization. In order to obtain an initial set of labels in an unsupervised fashion, we cluster the samples into different groups in the feature domain. We next propose to train Proportion-Support Vector Machine ($\propto$SVM) algorithm using label proportions rather than instance labels. The trained model is then employed to learn malignant-benign categorization of the tumors.

This paper is organized as follows. Section 2 describes related work pertaining to supervised and unsupervised learning for the diagnosis of lung nodules and IPMN. We present our MTL based supervised learning algorithm in Section 3. In Section 4, we introduce an unsupervised learning method adapted for the diagnosis of lung nodules and IPMN from CT and MRI scans, respectively. The experiments and results are discussed in Section 5. In the same section, we also study the influence of different deep learning features for unsupervised learning and establish an upper bound on the classification performance using supervised learning methods. Finally, Section 6 states discussions and concluding remarks.

\section{Related Work}
This section summarizes the advances in machine learning applied to medical imaging and CAD systems developed for lung cancer diagnosis. Since the automatic characterization of IPMN from MRI scans has not been extensively studied in the literature, relevant works are mostly selected from the clinical studies. Our work is the first in this regard.\\

\noindent\textbf{Imaging Features and Classifiers:} Conventionally, the risk stratification (classification) of lung nodules may require nodule segmentation, computation and selection of low-level features from the image, and the use of a  classifier/regressor. In the approach by~\cite{uchiyama2003quantitative}, different physical statistics including intensity measures were extracted and class labels were obtained using Artificial Neural Networks. In~\cite{el20113d}, lung nodules were segmented using appearance-based models followed by shape analysis using spherical harmonics. The last step involved $k$-nearest neighbor based classification. Another approach extended 2D texture features including Local Binary Patterns, Gabor and Haralick to 3D~\cite{han2015texture}. Classification using Support Vector Machine (SVM) was performed as the final step. In a different study, Way et al.~\cite{way2006computer}, implemented nodule segmentation via 3D active contours, and then applied rubber band straightening transform. A Linear Discriminant Analysis (LDA) classifier was applied to get class labels. Lee et al.~\cite{lee2010computer} introduced a feature selection based approach utilizing both clinical and imaging data. Information content and feature relevance were measured using an ensemble of genetic algorithm and random subspace method. Lastly, LDA was applied to obtain final classification on the condensed feature set.
In a recent work, spherical harmonics features were fused with deep learning features~\cite{buty2016characterization} and then RF classification was employed for lung nodule characterization. Hitherto, the application of CNN for nodule characterization has been limited to 2D space~\cite{chen2016bridging}, thus falling short of incorporating vital contextual and volumetric information. In another approach, Shin et al.~\cite{shen2015multi} employed CNN for the classification of lung nodules. Other than not completely 3D CNN, the approach didn't take into account high-level nodule attributes and required training an off-the-shelf classifier such as RF and SVM.

%\noindent\textbf{Image Attributes:} 
The information about different high-level image attributes had been found useful in the malignancy characterization of lung nodules. In a study exploring the correlation between malignancy and nodule attributes, \cite{furuya1999new} found that 82\% of the lobulated, 93\% of the ragged, 97\% of the densely spiculated, and 100\% of the halo nodules were malignant in a particular dataset. Automatic determination of lung nodule attributes and types had been explored in~\cite{ciompi2017towards}. The objective was to perform the classification of six different nodule types such as solid, non-solid, part-solid, calcified, perifissural and spiculated nodules. However, the approach is based on 2D CNN and fell short of estimating the malignancy of lung nodules. Furthermore, 66\% of the round nodules were determined as benign.

Although, not an objective in this paper, the detection of lung nodules has also been an active subject of interest among researchers~\cite{setio2016pulmonary,setio2017validation,khosravan2018s4nd}. A short but informative review of the most recent detection studies can be found in~\cite{khosravan2018s4nd}.\\ 

\noindent\textbf{Pancreatic Cysts (IPMN):} Although there has been considerable progress in developing automatic approaches to segment pancreas and its cysts~\cite{zhou2017deep,cai2016pancreas}, the use of advanced machine learning algorithms to perform fully automatic risk-stratification of IPMNs is limited. The approach by Hanania et al.~\cite{hanania2016quantitative} investigated the influence of 360 imaging features ranging from intensity, texture, and shape to stratify subjects as low or high-grade IPMN. In another example, Gazit et al.~\cite{gazit2017quantification} extracted texture and features from the solid component of segmented cysts followed by a feature selection and classification scheme. Both of these approaches~\cite{hanania2016quantitative,gazit2017quantification} required segmentation of cysts or pancreas and are evaluated on CT scans only.\\

\noindent\textbf{Unsupervised Learning:} Typically, the visual recognition and classification tasks are addressed using labeled data (supervision). However, for tasks where manually generating labels corresponding to large datasets is laborious and expensive, the use of unsupervised learning methods is of significant value. Unsupervised techniques had been used to solve problems in various domains ranging from object categorization~\cite{sivic2005discovering}, speech processing~\cite{kamper2015fully} and audio classification~\cite{lee2009unsupervised}. These methods conventionally relied on some complementary information provided with the data to improve learning, which may not be available for several classification tasks in medical imaging.

In medical imaging, there have been different approaches that used unsupervised learning for detection and diagnosis problems. The approach by Shin et al.~\cite{shin2013stacked} used stacked autoencoders for multiple organ detection in MRI scans. Vaidhya et al.~\cite{vaidhya2015multi} presented a brain tumor segmentation method with stacked denoising autoencoder evaluated on multi-sequence MRI images. In a work by Sivakumar et al.~\cite{sivakumar2012lung}, the segmentation of lung nodules is performed with unsupervised clustering methods. In another study, Kumar et al.~\cite{kumar2015lung} used features from autoencoder for lung nodule classification. These auto-encoder approaches, however, did not yield satisfactory classification results. Other than these, unsupervised deep learning has also been explored for mammographic risk prediction and breast density segmentation~\cite{kallenberg2016unsupervised}.

Unsupervised feature learning remains an active research area for the medical imaging community, more recently with Generative Adversarial Networks (GAN)~\cite{radford2015unsupervised}. In order to explore the information from unlabeled images, Zhang et al.~\cite{zhang2014ranking} described a semi-supervised method for the classification of four types of nodules. In sharp contrast to these approaches, the unsupervised learning strategies presented in this paper don't involve feature learning using auto-encoders. Using sets of hand-crafted as well as pre-trained deep learning features, we propose a new unsupervised learning algorithm where an initially estimated label set is progressively improved via proportion-SVM.\\

\begin{figure*}[t]
	\begin{subfigure}[c]{0.30\textwidth}
    \hspace{-0.25 cm}
    \centering
	\includegraphics[width=40 mm]{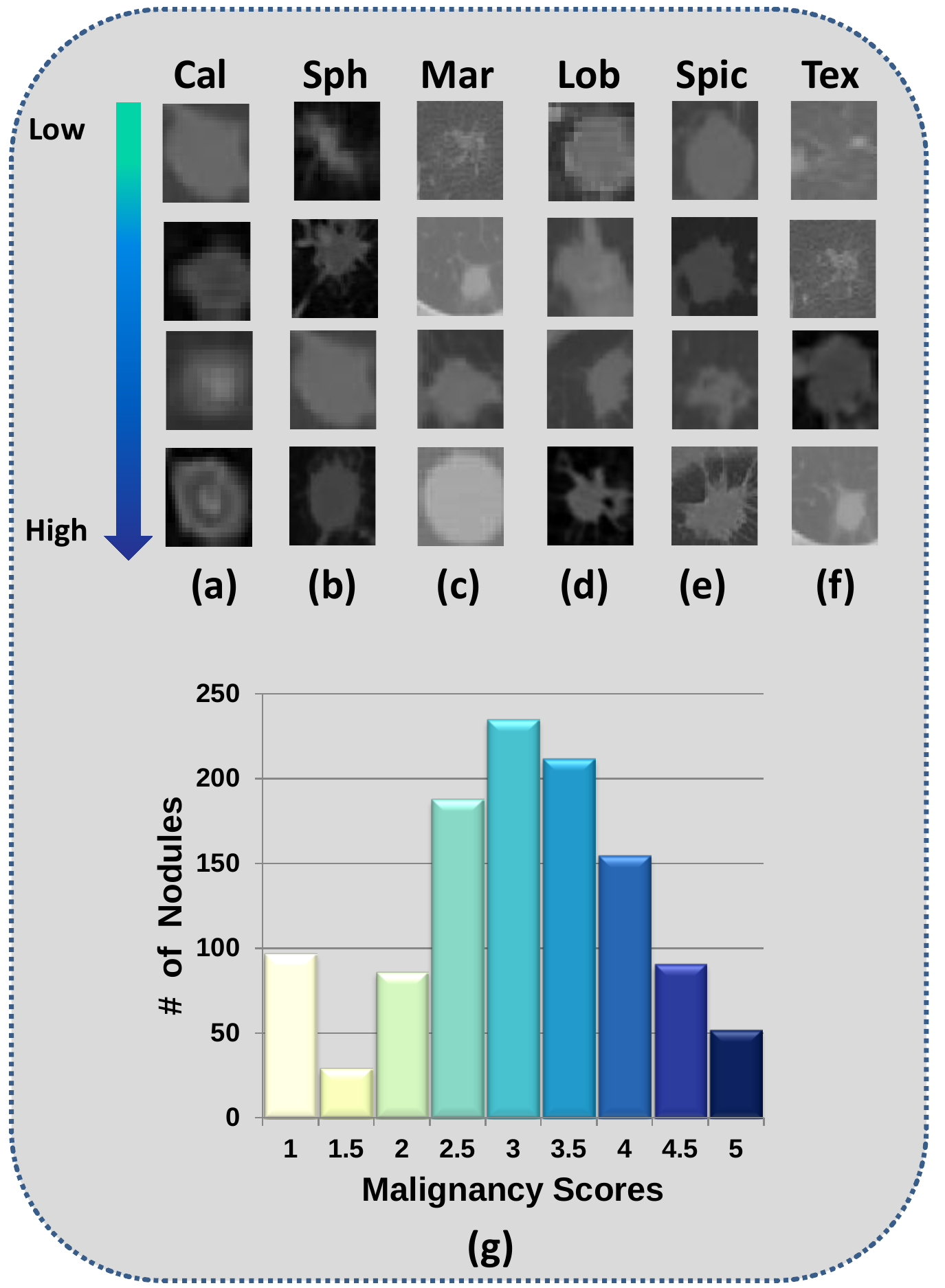}
    \caption{Lung nodule attributes}
    \end{subfigure}
    \begin{subfigure}[c]{0.75\textwidth}
    \centering
	\includegraphics[width=130 mm]{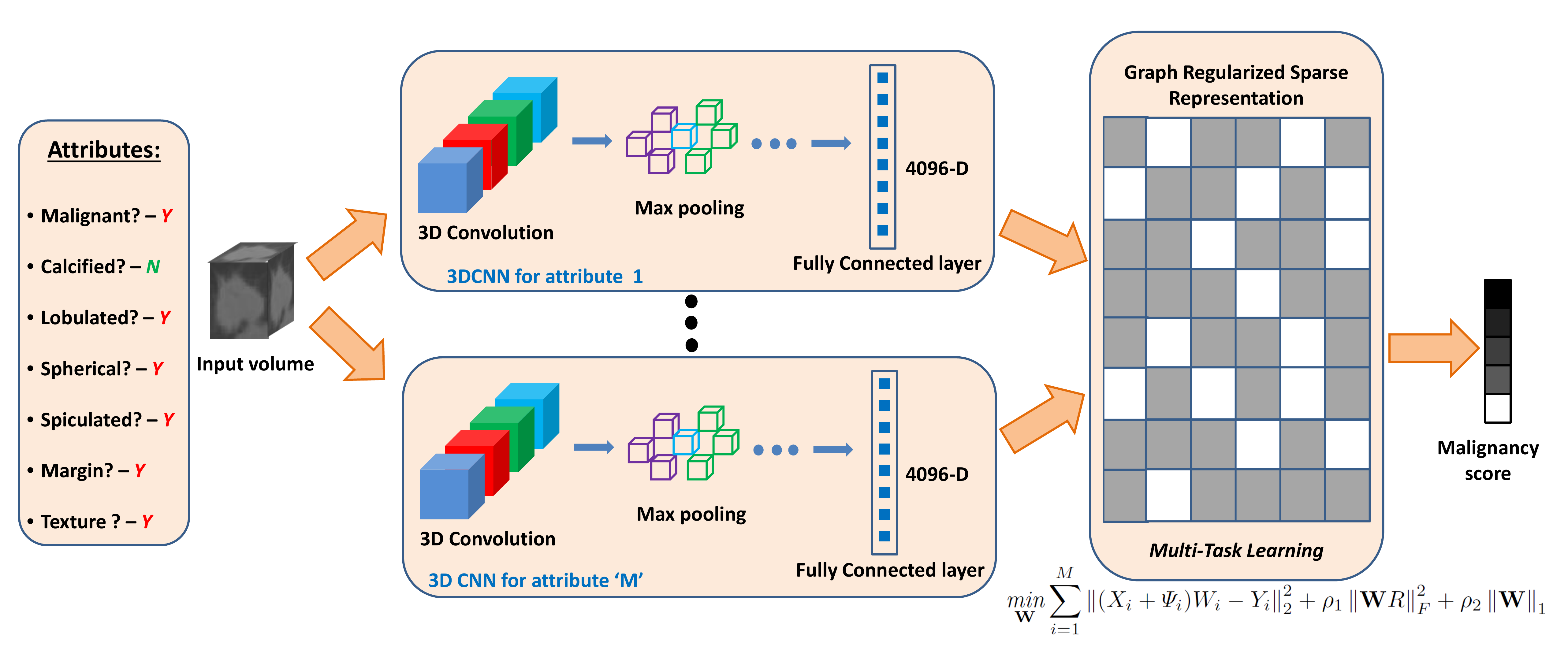}
    \caption{An overview of the proposed supervised approach}
    \end{subfigure}
    \caption{(A) A visualization of lung nodules having different levels of attributes. On moving from the top (attribute absent) to the bottom (attribute prominently visible), the prominence level of the attribute increase. Different attributes including calcification, sphericity, margin, lobulation, spiculation and texture can be seen in (a-f). The graph in (g) depicts the number of nodules with different malignancy levels in our experiments using the publicly available dataset~\cite{armato2011lung}. An overview of the proposed 3D CNN based graph regularized sparse MTL approach is presented in (B).}
  \label{fig:supworkflow}

\end{figure*}

\noindent \textbf{Our Contributions}\\
\noindent A block diagram representing different supervised and unsupervised schemes is presented in Figure~\ref{fig:blockdiag}. Overall, our main contributions in this work can be summarized as follows:

\begin{itemize}[topsep=1em,leftmargin=*]
\itemsep1em 
\item For lung nodule characterization, we present a 3D CNN based supervised learning approach to fully appreciate the anatomical information in 3D, which would be otherwise lost in the conventional 2D approaches. We use fine-tuning strategy to avoid the requirement for a large number of volumetric training examples for 3D CNN. For this purpose. we use a pre-trained network (which is trained on 1 million videos) and fine-tune it on the CT data. 

\item We introduce a graph regularized sparse MTL platform to integrate the complementary features from lung nodule attributes so as to improve malignancy prediction. Figure~\ref{fig:supworkflow}-A shows high-level lung nodule attributes having varying levels of prominence. 

\item We evaluate the proposed supervised and unsupervised learning algorithms to determine the characterization of lung nodules and IPMN cysts (Table I). In the era where the wave of deep learning has swept into almost all domains of visual analysis, we investigate the contribution of features extracted from different deep learning architectures. To the best of our knowledge, this is the first work to investigate the automatic diagnosis of IPMNs from MRI.

\item In the proposed unsupervised learning algorithm, instead of hard assigning labels, we estimate the label proportions in a data-driven manner. Additionally, to alleviate the effect of noisy labels (i.e. mislabels) obtained during clustering, we propose to employ $\propto$SVM, which is trained on label proportions only.
\end{itemize}

\section{Supervised Learning Methods}
\subsection{Problem Formulation}
Let $X=[x_1,x_2 \dots x_n]^T\in\mathbb{R}^{n \times d}$ represent the input features obtained from $n$ images of lung nodules each having a dimension $d$. %($\in\mathbb{R}^d$). 
Each data sample has an attribute/malignancy score given by $Y=[y_1,y_2 \dots y_n]$, where $Y^T\in \mathbb{R}^{n \times 1}$. While $X$ consists of features extracted from radiology images, and $Y$ represents the malignancy score over 1-5 scale where 1 represents benign and 5 represents malignant. In supervised learning, the labeled training data is used to learn the coefficient vector or the regression estimator $W\in\mathbb{R}^d$. In testing, $W$ is used to estimate $Y$ for an unseen feature/example. For regression, a regularizer is often added to prevent over-fitting. Hence, a classical least square regression turns into a \emph{constrained optimization} problem with $\ell_1$ regularization as:

\begin{equation}
\begin{split}
\minimum_{W}  \left \|XW-Y  \right \|_{2}^{2},
\text{ s.t. }   \left \| W \right \|_1 \leq t.
\end{split}
\label{eq:l1normc}
\end{equation}

In the above equation, the sparsity level of the coefficient vector $W=[w_1,w_2 \dots w_d]$ is controlled by a parameter $t$. Since the function in Eq.~(\ref{eq:l1normc}) is convex and the constraints define a convex set, a local minimizer of the objective function is subjected to constraints corresponding to a global minimizer. In the following subsections, we extend this supervised learning setting with deep learning and MTL concepts to characterize lung nodules as benign or malignant.

\subsection{3D Convolution Neural Network (CNN) and Fine-Tuning}
We use 3D CNN~\cite{tran2015learning} trained on Sports-1M dataset \cite{karpathy2014large} and fine-tune it on the lung nodule CT dataset. The Sports-1M dataset consists of 487 classes with 1 million videos. As the lung nodule dataset doesn't have a large number of training examples, fine-tuning is done to acquire dense feature representation from the Sports-1M. The 3D CNN architecture consists of 5 sets of convolution, 2 fully-connected and 1 soft-max classification layers. Each convolution set is followed by a max-pooling layer. The input to the 3D CNN comprises dimensions of 128x171x16, where 16 denotes the number of slices. Note that the images in the dataset are resized to have consistent dimensions such that the number of channels is 3 and the number of slices is fixed to 16. Hence, the overall input dimension can be considered as 3x16x128x171. The number of filters in the first 3 convolution layers are 64, 128 and 256 respectively, whereas there are 512 filters in the last 2 layers. The fully-connected layers have a dimension 4096 which is also the length of feature vectors used as an input to the MTL framework. Implementation details are mentioned in section V-C.

\subsection{Multi-task learning (MTL)}
Multi-task learning is an approach of learning multiple tasks simultaneously while considering disparities and similarities across those tasks. Given $M$  tasks, the goal is to improve the learning of a model for task $i$, ($i\in M$) by using the information contained in the $M$ tasks. We formulate the malignancy prediction of lung nodules as an MTL problem, where visual attributes of lung nodules are considered as distinct tasks (Figure~\ref{fig:supworkflow}A). In a typical MTL problem, initially, the correlation between $M$ tasks and the shared feature representations are not known. The aim in the MTL approach is to learn a joint model while exploiting the dependencies among visual attributes (tasks) in feature space. In other words, we utilize visual attributes and exploit their feature level dependencies so as to improve regressing malignancy using other attributes.

As shown in Figure~\ref{fig:supworkflow}B, we design lung tumor characterization as an MTL problem, where each task has model parameters $W_i$, which are utilized to characterize the corresponding task $i$. When $\textbf{W}$ $=[W_1,W_2 \dots W_M] \in\mathbb{R}^{d \times M}$ constitutes a rectangular matrix, rank can be considered as a natural extension to cardinality, and nuclear/trace norm leads to low rank solutions. In some cases nuclear norm regularization can be considered as the $\ell_1$-norm of the singular values~\cite{recht2010guaranteed}. Trace norm, the sum of singular values, is the convex envelope of the rank of a matrix (which is non-convex), where the matrices are considered on a unit ball. After substituting, $\ell_1$-norm by trace norm, the least square loss function with trace norm regularization can be formulated as:

\begin{equation}
\minimum_{\mathbf{W}}\sum_{i=1}^{M} \left \|\mathbf{X_i}W_i-\mathbf{Y_i}  \right \|_{2}^{2}+\rho\left \| \mathbf{W} \right \|_{*},
\label{eq:mtltrace}
\end{equation}

\noindent where $\rho$ adjusts the rank of the matrix \textbf{W}, and $\left \| \mathbf{W} \right \|_{*}=\sum_{i=1}\sigma_i(\mathbf{W})$ is the trace-norm where $\sigma$ denotes singular values. However, as in trace-norm, the assumption about models sharing a common subspace is restrictive for some applications.

As the task relationships are often unknown and are learned from data, we represent tasks and their relations in the form of a graph. Let $\Upsilon=(V,E)$ represent a complete graph in which nodes $V$ correspond to the tasks and the edges $E$ model any affinity between the tasks. In such case, a regularization can be applied on the graph modeling task dependencies~\cite{zhou2011malsar}. The complete graph can be modeled as a structure matrix $S=[e^1,e^2 \dots e^{\left \| E \right \|}]\in\mathbb{R}^{M\times\left \| E \right \|}$ where the deviation between the pairs of tasks can be regularized as:

\begin{equation}
\left \|\mathbf{W}S  \right \|_{F}^{2}=\sum_{i=1}^{\left \| E \right \|} \left \|\mathbf{W}e^i \right \|_{2}^{2}=
\sum_{i=1}^{\left \| E \right \|} \left \|\mathbf{W}_{e^{i}_{a}}-\mathbf{W}_{e^{i}_{b}} \right \|_{2}^{2},
\label{eq:mtltrace2}
\end{equation}
\noindent here, $e^{i}_{a}$, $e^{i}_{b}$ are the edges between the nodes $a$ and $b$, and $e^{i}\in\mathbb{R}^{M}$. The matrix $S$ defines an incidence matrix where $e^{i}_{a}$ and $e^{i}_{b}$ are assigned to 1 and -1, respectively, if nodes $a$ and $b$ are connected in the graph. Eq.~(\ref{eq:mtltrace2}) can be further explained as:

\begin{equation}
\left \|\mathbf{W}S  \right \|_{F}^{2}=\text{tr}((\mathbf{W}S)^T (\mathbf{W}S))=\text{tr}(\mathbf{W}SS^T \mathbf{W}^T)=\text{tr}(\mathbf{W} \mathcal{L}\mathbf{W}^T),
\label{eq:mtltrace}
\end{equation}

\noindent where $\mathcal{L}=SS^T$ is the Laplacian matrix and `tr' represents the trace of a matrix. The method to compute structure matrix $S$ is discussed in Section V-C.

The malignancy prediction equation can be further regularized because there are still other uncertainties to consider, i.e., disagreement between radiologists' visual interpretations for a given nodule. For instance, while one radiologist may give a malignancy score of $x_1^j$ for a nodule $j$, the other may give a score of $x_2^j$ for the same nodule. In order to reflect these uncertainties in our algorithm, we formulate a scoring function which models such inconsistencies:

\begin{equation}
\Psi(j) =\left (\exp(\frac{-\sum_{r}(x_r^j-\mu^j )^{2}}{2\sigma^j} )\right )^{-1}.
\end{equation}

For a particular example $j$, this inconsistency measure can be represented as $\Psi(j)$. $x_r^j$ is the score given by the $r^{th}$ radiologist (expert) whereas $\mu^j$ and $\sigma^j$ represent mean and standard deviation of the scores, respectively. We calculate this inconsistency score for all the tasks under consideration and for simplicity we have omitted the index for the tasks. The final objective function of graph regularized sparse least square optimization with the inconsistency measure can expressed as:
\begin{equation}
\minimum_{\mathbf{W}}\sum_{i=1}^{M}\overset{{\textcircled{1}}}{\overbrace{\left \|(\mathbf{X_i}+\mathbf{\Psi_i})W_i-\mathbf{Y_i}  \right \|_{2}^{2}}}+\overset{{\textcircled{2}}}{\overbrace{\rho_1\left \| \mathbf{W}S \right \|_{F}^{2}}}+\overset{{\textcircled{3}}}{\overbrace{\rho_2\left \| \mathbf{W} \right \|_{1}}},
\label{eq:finaleq}
\end{equation}

%+\rho_3\left \| \mathbf{W} \right \|_{F}^{2}

\noindent where $\rho_1$ tunes the penalty degree for graph structure and $\rho_2$ handles the sparsity level. In Eq.~(\ref{eq:finaleq}), the least square loss function \textcircled{1} observes decoupling of tasks whereas \textcircled{2} and \textcircled{3} model their interdependencies, so as to learn joint representation.

\subsection{Optimization}
In order to solve Eq.~(\ref{eq:finaleq}), the conventional approach is to use standard gradient descent as an optimization algorithm. However, standard gradient descent cannot be applied here because the $\ell_1-$norm is not differentiable at $\mathbf{W}=0$ and gradient descent approach fails to provide sparse solutions~\cite{shalev2011stochastic}. Since the optimization function in the above equation has both smooth and non-smooth convex parts, it can be solved after replacing the non-smooth part with its estimates. In other words, the $\ell_1$-norm in the above equation is the non-smooth part and the proximal operator can be used for its estimation. For this purpose, we utilize \emph{accelerated proximal gradient method} \cite{nesterov2013introductory}, the first order gradient method having a convergence rate of $O(1/m^2)$, where $m$ controls the number of iterations. %Note that in 

\section{Unsupervised Learning Methods}
\begin{figure*}[t]
\centering
\includegraphics[width=140 mm]{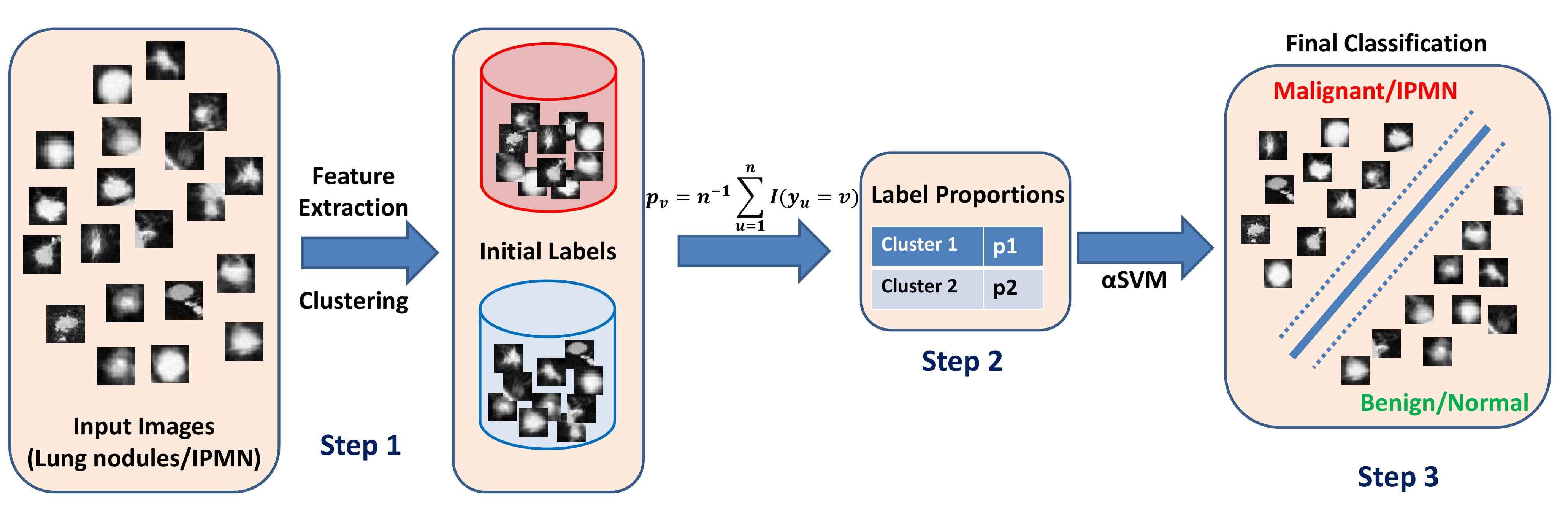}
\caption{An outline of the proposed unsupervised approach. Given the input images, we compute GIST features and perform $k$-means clustering to get the initial set of labels which can be noisy. Using the set of labels, we compute label proportions corresponding to each cluster/group (Eq. (9)). We finally employ $\propto$SVM to learn a discriminative model using the features and label proportions.}
\label{fig:workflow}
%\vspace{-0.3 cm}
\end{figure*}

Since annotating medical images is laborious, expensive and time-consuming, in the second part of this paper, we explore the potential of unsupervised learning approaches for tumor characterization problems. As illustrated in Figure~\ref{fig:workflow}, our proposed unsupervised framework includes three steps. First, we perform clustering on the appearance features obtained from the images to estimate an initial set of labels. Then, using the obtained initial labels, we compute label proportions corresponding to each cluster. Finally, we use the initial cluster assignments and label proportions to learn the categorization of tumors.

\begin{table*}[t]
\caption{List and details of different experiments performed for supervised and unsupervised learning along with their evaluation sets.}
\centering
  \begin{tabular}{|P{2.0cm}|P{5.0cm}|P{2.5cm}|}
\hline
\textbf{Experiments} & \textbf{Details} & \textbf{Evaluation Set} \\ \hline
E1 & Supervised learning,
3D CNN based Multi-task learning with attributes, fine-tuning (C3D) network
  & 
3D dataset: Malignancy score regression of Lung nodules (CT) \\ \hline
E2 & 
Unsupervised learning, GIST features, Proportion-SVM
  & \multirow{7}{7em}{\centering 2D dataset: Lung nodules (CT) and 
  IPMN classification (MRI)} \\ \cline{1-2}
E3 &  
Unsupervised learning, features from different layers of 2D VGG network
 & \\ \cline{1-2}
E4 &  
Supervised learning to establish classification upper-bound, GIST and VGG features with SVM and RF
 & \\ \hline
% - & -  & \multirow{2}{*}{Note 3} \\ \cline{1-2}
% - & V & \\ \hline
\end{tabular}
\label{tab:exp}
\end{table*}

\subsection{Initial Label Estimation}
Let $X=[x_1,x_2 \dots x_n]^T\in\mathbb{R}^{n \times d}$ represent the input matrix which contains features from $n$ images such that $x \in \mathbb{R}^d$. We then cluster the data into $2\leq k<n$ clusters using $k$-means algorithm. Let $A$ represent $|X| \times k$ assignment matrix which denotes the membership assignment of each sample to a cluster. The optimal clustering would minimize the following objective function:
\begin{equation}
\begin{split}
\argminB_{\mu_v,A} \sum_{v=1}^{k}A(u,v)\left \|x_u-\mu_v  \right \|^2,\\
\text{ s.t. }   A(u,v)=0 \vee 1,\sum_vA(u,v)=1 
\end{split}
\label{eq:kmeans}
\end{equation}

\noindent where $\mu_v$ is the mean of the samples in cluster $v$. The assignment matrix $A$ can then be used to estimate labels $c$. These labels are only used for estimating label proportions of the clustered data for the purpose of training a new algorithm which we adapt for our problem, i.e., proportion-SVM ($\propto$SVM). The rationale behind this proportion comes from the clustering notion where data is divided into groups/clusters and each cluster corresponds to a particular class. In our work, specifically, clustering is only an initial step to estimate cluster assignments that are progressively refined in the subsequent steps.%Particularly when the class number is known (i.e., 2 groups), then clustering can be used for classification.}

\subsection{Learning with the Estimated Labels}

Since our initial label estimation approach is unsupervised, there are uncertainties associated with them. It is, therefore, reasonable to assume that learning a discriminative model based on these noisy instance level labels can deteriorate classification performance. In order to address this issue, we model the instance level labels as \textit{latent} variables and thereby consider group/bag level labels.

Inspired by $\propto$SVM approach~\cite{yu2013propto}, which models the latent instance level variables using the known group level label proportions, we formulate our learning problem such that clusters are analogous to the groups. In our formulation, each cluster $v$ can be represented as a group such that the majority of samples belong to the class $v$. Considering the groups to be disjoint such that $\bigcup_{v=1}^{k}$\textbf{$\Omega_v$}$={1,2,\dots n}$ , and $\Omega$ represents groups; the objective function of the large-margin $\propto$SVM after convex relaxation can be formulated as:

\begin{equation}
\begin{split}
\minimum_{\mathbf{c} \in \mathcal{C}} \minimum_{w} \left(\frac{1}{2}w^Tw+K\sum_{u=1}^{n}L(c_u,w^T\phi(x))\right)\\
\mathcal{C}=\biggl\{ \mathbf{c} \bigg| \left | \widetilde{p_v}(\mathbf{c})-p_v \right | \leq \epsilon ,c_u \in \left \{ -1,1 \right \} \forall_{v=1}^{k} \biggr\},
\end{split}
\label{eq:propsvm}
\end{equation}

\noindent where $\widetilde{p}$ and $p$ represent the estimated and true label proportions, respectively. In Eq.~(\ref{eq:propsvm}), $\mathbf{c}$ is the set of instance level labels, $\phi(.)$ is the input feature, $K$ denotes cost parameter and $L(.)$ represents the hinge-loss function for maximum-margin classifiers such as SVM. An alternative approach based on training a standard SVM classifier with clustering assignments is discussed in Section V-D.

The optimization in Eq.~(\ref{eq:propsvm}) is, in fact, an instance of Multiple Kernel Learning, which can be solved using the cutting plane method where the set of active constraints is incrementally computed. The goal is to find the most violated constraint, however, the objective function still decreases even by further relaxation and aiming for any violated constraint. Further details about optimization can be studied in~\cite{yu2013propto}.

\begin{figure*}[t]
\centering
\includegraphics[width=170 mm]{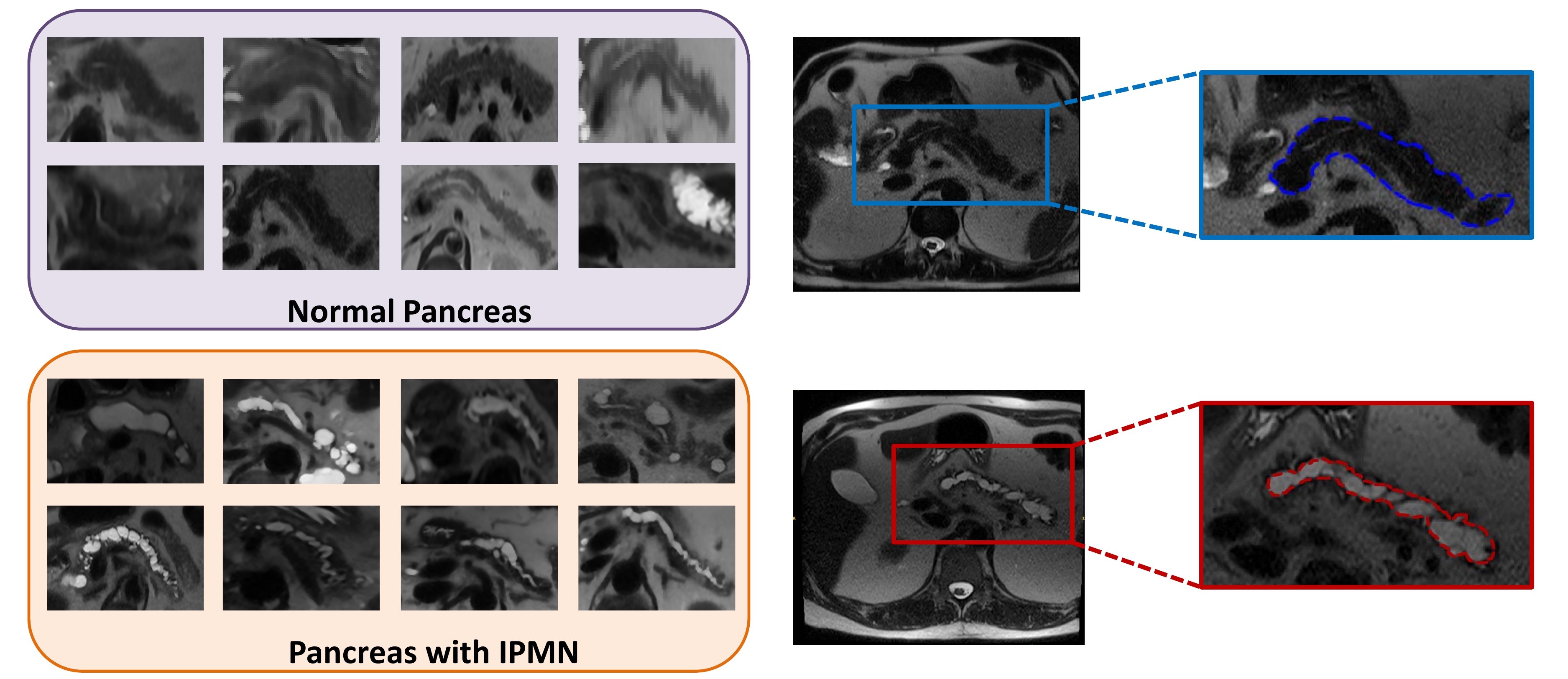}
\caption{Axial T2 MRI scans illustrating pancreas. The top row shows different ROIs of pancreas, along with a magnified view of a normal pancreas (outlined in blue). The bottom row shows ROIs from subjects with IPMN in the pancreas, which is outlined in red.}
\label{fig:IPMN}
%\vspace{-0.3 cm}
\end{figure*}

\subsection{Calculating Label Proportions}
In the conventional $\propto$SVM approach, the label proportions are known a priori. Since our approach is unsupervised, both instance level labels and group label proportions are unknown. Moreover, establishing strong assumptions about the label proportions may affect learning. It is, however, reasonable to assume that a large number of instances in any group carry the same label and there may be a small number of instances which are outliers. The label proportions serve as a soft-label for a bag where a bag can be considered as a super-instance. In order to determine the label proportions in a data-driven manner, we use the estimated labels obtained from clustering. The label proportion $p_v$ corresponding to the group $v$ can be represented as:
\begin{equation}
p_{v}=n^{-1} \sum_{u=1}^{n}I(y_u=v),
\label{eq:label}
\end{equation}

\noindent where I(.) is the indicator function which yields 1 when $y_u=v$. The $\propto$SVM is trained using the image features and label proportions to classify the testing data. It is important to mention that the ground truth labels (benign/malignant labels) are used only to evaluate the proposed framework and are not used in estimating label proportions or training of the proportion-SVM. In addition, clustering and label proportion calculation are only performed on the training data and the testing data remains completely unseen for $\propto$SVM. The number of clusters is fixed as 2, i.e. benign and malignant classes and the result was checked to assign benign and malignant labels to the clusters.
 
\section{Experiments}

\subsection{Data for Lung Nodules}
For test and evaluation, we used LIDC-IDRI dataset from Lung Image Database Consortium \cite{armato2011lung}, which is one of the largest publicly available lung nodule dataset. The dataset comprises 1018 CT scans with a slice thickness varying from 0.45 mm to 5.0 mm. At most four radiologists annotated those lung nodules which have diameters equal to or greater than 3.0 mm.

We considered nodules which were interpreted by at least three radiologists for evaluations. The number of nodules fulfilling this criterion was 1340. As a nodule may have different malignancy and attribute scores provided by different radiologists, their mean scores were used. The nodules have scores corresponding to these six attributes: (i) calcification, (ii) lobulation, (iii) spiculation, (iv) sphericity, (v) margin and (vi) texture as well as malignancy (Figure~\ref{fig:supworkflow}). The malignancy scores ranged from 1 to 5 where 1 denoted benign and 5 meant highly malignant nodules. To account for malignancy indecision among radiologists, we excluded nodules with a mean score of 3. The final evaluation set included 509 malignant and 635 benign nodules. As a pre-processing step, the images were resampled to be isotropic so as to have 0.5 mm spacing in each dimension.

\subsection{Data for IPMN}
The data for the classification of IPMN contains T2 MRI axial scans from 171 subjects. The scans were labeled by a radiologist as normal or IPMN. Out of 171 scans, 38 subjects were normal, whereas the rest of 133 were from subjects diagnosed with IPMN. The in-plane spacing (xy-plane) of the scan was ranging from 0.468 mm to 1.406 mm. As pre-processing, we first employ N4 bias field correction~\cite{tustison2010n4itk} to each image in order to normalize variations in image intensity. We then apply curvature anisotropic image filter to smooth image while preserving edges. For experiments, 2D axial slices with pancreas (and IPMN) are cropped to generate Region of Interest (ROI) as shown in Figure~\ref{fig:IPMN}. The large intra-class variation, especially due to varying shapes of the pancreas can also be observed in Figure~\ref{fig:IPMN}. A list of different supervised and unsupervised learning experiments along with their evaluation sets is tabulated in Table~\ref{tab:exp}.

\begin{table*}[t]

\begin{center}
\caption{The comparison of the proposed approach with other methods using regression accuracy and mean absolute score difference for lung nodule characterization.}
\label{table:Results}%\hspace{-0.1in}
\label{table:quan_acc}
\normalsize{
\begin{tabular}{l@{\hspace{0.8in}}c@{\hspace{0.4in}}c@{\hspace{0.4in}}c}
\toprule[1.5pt] \multirow{2}{*}{\textbf{Methods}}   & \multirow{2}{*}{\textbf{Accuracy}} & \multirow{2}{*}{\textbf{Mean Score}} \\ 
\multirow{2}{*}{}   & \multirow{2}{*}{\textbf{\%}} & \multirow{2}{*}{\textbf{Difference}} \\
\\
\cmidrule(r){1-4}
GIST features + LASSO    &      76.83     &      0.675   &        \\
GIST features + RR   &     76.48   &      0.674  &        \\
3D CNN features + LASSO (\textit{Pre-trained})   &     86.02     &     0.530  &        \\
3D CNN features + RR (\textit{Pre-trained})    &      82.00    &      0.597   &        \\
3D CNN features + LASSO (\textit{Fine-tuned})    &      88.04    &      0.497  &        \\
3D CNN features + RR (\textit{Fine-tuned})    &      84.53     &      0.550   &        \\
3D CNN MTL with Trace norm &      80.08     &      0.626  &        \\
\textbf{Proposed (3D CNN with Multi-task Learning- Eq. 7)}     &     \textbf{91.26}    &      \textbf{0.459}   &        \\
\toprule[1.5pt]
\end{tabular}
}
\end{center}
%\vspace{-0.2 in}
\end{table*}

\subsection{Evaluation and Results- Supervised Learning}
We fine-tuned the 3D CNN network trained on Sports-1M dataset \cite{karpathy2014large} which had 487 classes. In order to train the network with binary labels for malignancy and the six attributes we used the mid-point as a pivot and labeled samples as positive (or negative) based on their scores being greater (or lesser) than the pivot. In our context, malignancy and attributes are characterized as tasks. The C3D was fine-tuned with these 7 tasks and 10 fold cross-validation was conducted. The requirement to have a large amount of labeled training data was evaded by fine-tuning the network. Since the input to the network required 3 channel image sequences with at least 16 slices, we concatenated the gray level axial channel as the other two channels.

Additionally, in order to ascertain that all input volumes have 16 slices, we performed interpolation where warranted. The final feature representation was obtained from the first fully connected layer of 3D CNN consisting of 4096-dimensions. 

For computing structure matrix $S$, we calculate the correlation between different tasks by estimating the normalized coefficient matrix $\mathbf{W}$ via least square loss function with lasso followed by the calculation of correlation coefficient matrix \cite{zhou2011malsar}. In order to get a binary graph structure matrix, we thresholded the correlation coefficient matrix. As priors in Eq.~(\ref{eq:finaleq}) we used $\rho_1$ and $\rho_2$ as 1 and 10 respectively. Finally, to obtain the malignancy score for test images, the features from the network trained on malignancy were multiplied with the corresponding task coefficient vector $W$.

We evaluated our proposed approach using both classification and regression metrics. For classification, we considered a nodule to be successfully classified if its predicted score lies in $\pm$1 of the ground truth score. For regression, we calculated average absolute score difference between the predicted score and the true score. The comparison of our proposed MTL approach with approaches including GIST features~\cite{GIST}, 3D CNN features from pre-trained network + LASSO, Ridge Regression (RR) and 3D CNN MTL+trace norm is tabulated in Table \ref{table:quan_acc}. It can be observed that our proposed graph regularized MTL performs significantly better than other approaches both in terms of classification accuracy as well as the mean score difference. The gain in classification accuracy was found to be 15\% and 11\% for GIST and trace-norm respectively. In comparison with the pre-trained network, we obtain an improvement of 5\% with proposed MTL. In addition, our proposed approach reduces the average absolute score difference for GIST by 32\% and for trace-norm by 27\%.

\subsection{Evaluations and Results- Unsupervised Learning}
For unsupervised learning, evaluations were performed on both lung nodules and IPMN datasets. In order to compute image level features, we used GIST descriptors~\cite{GIST}. The number of clusters is fixed as 2, which accounts for benign and malignant classes. The clustering result was checked to assign benign and malignant labels to the clusters. We used 10 fold cross-validation to evaluate our proposed approach. The training samples along with the label proportions generated using clustering served as the input to $\propto$SVM with a linear kernel.  

\begin{table*}[h!]
\begin{center}
%\hspace{-3.0in}
\caption{Average classification accuracy, sensitivity, and specificity of the proposed \textit{unsupervised} approach for IPMN and lung nodule classification with other methods}
\small{
\begin{tabular}{l@{\hspace{0.01in}}c@{\hspace{0.15in}}c@{\hspace{0.08in}}c@{\hspace{0.08in}}c@{\hspace{0.08in}}c}
 \hline\hline \multirow{2}{*}{\textbf{Evaluation Set}} & \multirow{2}{*}{\textbf{Methods}} & \multirow{2}{*}{\textbf{Accuracy}} & \multirow{2}{*}{\textbf{Sensitivity}} & \multirow{2}{*}{\textbf{Specificity}} \\ \\
\cmidrule(r){1-6}
 \multirow{4}{8em}{\textit{IPMN Classification}}   & Clustering    &   49.18\% &	45.34\%	& 62.83\% & \\
   & Clustering + RF    &   53.20\% &	51.28\%	&  \textbf{69.33\%} & \\
                          & Clustering + SVM &   52.03\%  &	51.96\% &	50.5\% & \\
                           & \textbf{Proposed approach}     &     \textbf{58.04\%} &	\textbf{58.61\%} &	41.67\% &  \\
                            \cmidrule(r){2-6}
 \multirow{4}{10em}{\textit{Lung Nodule Classification}}   & Clustering    &      54.83\%     &      48.69\%   & 60.04\%   &      \\
 & Clustering + RF    &   76.74\% &	58.59\%	&  \textbf{91.40\%} & \\
                          & Clustering + SVM &      76.04\%     &      57.08\%  &   91.28\% &     \\
                           & \textbf{Proposed approach}     &     \textbf{78.06\%}    &      \textbf{77.85\%}   & 78.28\% &       \\ [1ex] % [1ex] adds vertical space
\hline \hline
\end{tabular}
}
\label{table:unsupervised}
\end{center}
\end{table*}

To evaluate our unsupervised approach we used accuracy, sensitivity and specificity as metrics. It can be observed in Table~\ref{table:unsupervised} that the proposed combination of clustering and $\propto$SVM significantly outperforms other approaches in accuracy and sensitivity. In comparison with clustering+SVM, the proposed framework yields almost 21\% improvement in sensitivity for lung nodules and around 7\% improvement for IPMN classification. The low sensitivity and high specificity of clustering, clustering+SVM, and clustering+RF approaches can be explained by disproportionate assignment of instances as benign (normal) by these approaches, which is not found in the proposed approach. At the same time, the proposed approach records around 24\% and 9\% improvement in accuracy as compared to clustering for lung nodules and IPMN, respectively. \\

\begin{figure*}[t]
\centering
\includegraphics[width=\textwidth]{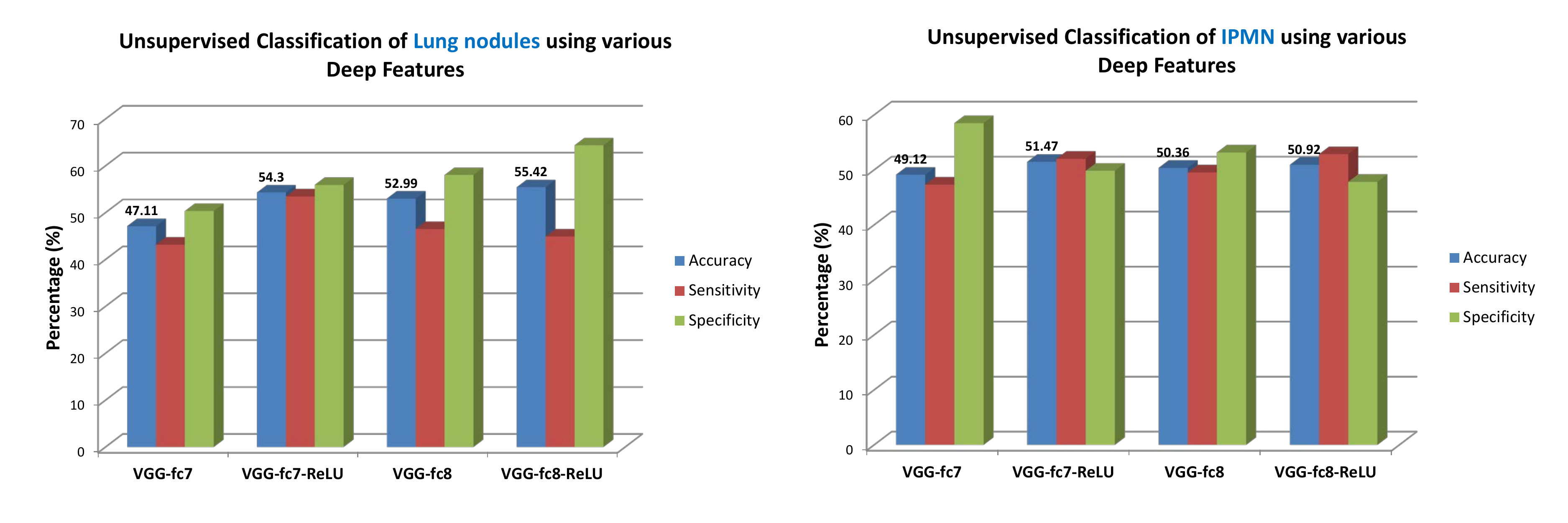}
\caption{Influence of deep learning features obtained from different layers of a VGG network with and without ReLU non-linearities. The graph on the left shows accuracy, sensitivity and specificity for unsupervised lung nodule classification (clustering), whereas the right one shows the corresponding results for IPMN.}
\label{fig:deepfeats}
%\vspace{-0.3 cm}
\end{figure*}

\noindent\textbf{Are Deep Features good for Unsupervised Classification?}\\
\noindent Given the success of deep learning features for image classification and their popularity with the medical imaging community, we explored their performance to classify lung nodules and IPMN in an unsupervised manner. For this purpose, we used a pre-trained deep CNN architecture to extract features and then perform clustering to obtain baseline classification performance. We extracted features from fully-connected layers 7 and 8 of Fast-VGG~\cite{VGG} with and without applying ReLU non-linearity. Classification accuracy, using clustering over these features is shown in Figure~\ref{fig:deepfeats}.

It can be seen in Figure~\ref{fig:deepfeats} that the features with non-linearity (ReLU) are more discriminative for classification using clustering as compared to without ReLU. The same trend can be observed for both lung nodules and IPMN classification using VGG-fc7 and VGG-fc8 layers. Owing to the larger evaluation set, the influence of ReLU is more prominent for lung nodules as compared to IPMN. Although the results between VGG-fc7 and VGG-fc8 are not substantially different, the highest accuracy for IPMN can be obtained by using VGG-fc7-ReLU features and for lung nodules by using VGG-fc8-ReLU features. The non-linearity induced by ReLU clips the negative values to zero, which can sparsify the feature vector and can reduce overfitting. Additionally, it can be seen that GIST features yield comparable performance than deep features (Table~\ref{table:unsupervised}). This can be explained by the fact that the deep networks were trained on ImageNet dataset so the filters in the networks were more tuned to the variations in natural images than medical images. Classification improvement can be expected with unsupervised feature learning techniques such as GANs~\cite{radford2015unsupervised}.\\

\begin{table*}[t]
\begin{center}
\caption{Classification of IPMN and Lung Nodules using different features and supervised learning classifiers.}
\small{
\begin{tabular}%{l*{7}{c}}
{l@{\hspace{0.2in}}c@{\hspace{0.3in}}c@{\hspace{0.2in}}c@{\hspace{0.2in}}c@{\hspace{0.2in}}c@{\hspace{0.1in}}c}

 \hline\hline \multirow{2}{*}{\textbf{Evaluation Set}} & \multirow{2}{*}{\textbf{Features}} & \multirow{2}{*}{\textbf{Classifiers}} & \multirow{2}{*}{\textbf{Accuracy (\%)}} & \multirow{2}{*}{\textbf{Sensitivity (\%)}} & \multirow{2}{*}{\textbf{Specificity (\%)}} \\ \\
\cmidrule(r){1-6}

 \multirow{6}{8em}{\textit{IPMN Classification}} & \multirow{2}{2em}{GIST} &  {SVM} &  {76.05} &  {83.65} &  \textbf{52.67}   \\
                          & & RF & 81.9 &  93.69 &   43.0 \\
                          \cline{2-6}
                          & \multirow{2}{4em}{VGG-fc7} &  {SVM} &  {84.18} &  {96.91} &  {44.83}   \\
                          & & RF & 81.96  &  94.61 &   42.83 \\
                          
                          \cline{2-6}
                         & \multirow{2}{4em}{VGG-fc8} &  {SVM} &  \textbf{84.22} &  \textbf{97.2} &  {46.5}   \\
                          & & RF & 80.82  &  93.4 &   45.67 \\
 \hline
  \multirow{6}{8em}{\textit{Lung Nodule Classification}} & \multirow{2}{2em}{GIST} &  {SVM} &  {81.56} &  {71.31} &  {90.02}   \\
                          & & RF & 81.64  &  76.47 &   \textbf{85.97} \\
                          \cline{2-6}
                          & \multirow{2}{4em}{VGG-fc7} &  {SVM} &  {77.97} &  {75.2} &  {80.6}   \\
                          & & RF & \textbf{81.73} &  \textbf{78.24} &   84.59 \\
                          
                          \cline{2-6}
                         & \multirow{2}{4em}{VGG-fc8} &  {SVM} &  {78.76} &  {74.67} &  {82.29}   \\
                          & & RF & 80.51  &  76.03 &   84.24 \\
\hline
 \hline
\end{tabular}
}
\label{table:supervised}
\end{center}
\end{table*}

\noindent\textbf{Classification using Supervised Learning}\\
\noindent In order to establish the upper-bound on the classification performance, we trained linear SVM and Random Forest using GIST and different deep learning features with ground truth labels on the same 10 fold cross-validations sets. Table~\ref{table:supervised} lists the classification accuracy, sensitivity, and specificity using GIST, VGG-fc7 and VGG-fc8 features for both IPMN and lung nodules. For both VGG-fc7 and VGG-fc8, we used features after ReLU since they are found to be more discriminative (Figure~\ref{fig:deepfeats}). Interestingly, for lung nodules, VGG-fc7 features along with RF classifier are reported to have comparable results to the combination of GIST and RF classifier. This can be explained by the fact that deep networks are pre-trained on ImageNet dataset as compared to handcrafted features such as GIST, which don't require any training. On the other hand, for smaller datasets such as IPMN, deep features are found to perform better as compared to GIST. In order to balance the number of positive (IPMN) and negative (normal) examples, which can be a critical drawback otherwise, we performed Adaptive Synthetic Sampling~\cite{he2008adasyn}. This was done to generate synthetic examples in terms of features from the minority class (normal).\\

\section{Discussion and Concluding Remarks}
In this study, we present a framework for the malignancy determination of lung nodules with 3D CNN based graph regularized sparse MTL. To the best of our knowledge, this is the first work where MTL and transfer learning are studied for 3D deep networks to improve risk stratification of lung nodules. Usually, the data sharing for medical imaging is highly regulated and the accessibility of experts (radiologists) to label these images is limited. As a consequence, the access to the crowdsourced and publicly gathered and annotated data such as videos may help in obtaining discriminative features for medical image analysis. 

We also analyzed the significance of different imaging attributes corresponding to lung nodules including spiculation, texture, calcification and others for risk assessment. Instead of manually modeling these attributes we utilized 3D CNN to learn rich feature representations associated with these attributes. The graph regularized sparse MTL framework was employed to integrate 3D CNN features from these attributes. We have found the features associated with these attributes complementary to those corresponding to malignancy.

In the second part of this study, we explored the potential of unsupervised learning for malignancy determination. Since in most medical imaging tasks radiologists are required to get annotations, acquiring labels to learn machine learning models is more cumbersome and expensive as compared to other computer vision tasks. In order to address this challenge, we employed clustering to obtain an initial set of labels and progressively refined them with $\propto$SVM. We obtained promising results and our proposed approach outperformed the other methods in evaluation metrics.

Following up on the application of deep learning for almost all tasks in the visual domain, we studied the influence of different pre-trained deep networks for lung nodule classification. For some instances, we found that commonly used imaging features such as GIST have comparable results as those obtained from pre-trained network features. This observation can be explained by the fact that the deep networks were trained on ImageNet classification tasks so the filters in CNN were more tuned to the nuances in natural images as compared to medical images.

To the best of our knowledge, this is one of the first and the largest evaluation of a CAD system for IPMN classification. CAD systems for IPMN classification are relatively newer research problems and there is a need to explore the use of different imaging modalities to improve classification. Although MRI remains the most common modality to study pancreatic cysts, CT images can also be used as a complementary imaging modality due to its higher resolution and its ability to capture smaller cysts. Additionally, a combination of T2-weighted, contrast-enhanced and unenhanced T1-weighted sequences can help improve detection and diagnosis of IPMN~\cite{kalb2009mr}. In this regard, multi-modal deep learning architectures can be deemed useful~\cite{ma2015multimodal}.
The detection and segmentation of pancreas can also be useful to make a better prediction about the presence of IPMN and cysts. Due to its anatomy, the pancreas is a challenging
organ to segment, particularly in MRI images. To address this challenge, other imaging modalities can be utilized for joint segmentation and diagnosis of pancreatic cysts and IPMN. Furthermore, visualization of activation maps can be quite useful for the clinicians to identify new imaging biomarkers that can be employed for diagnosis in the future.

The future prospects of using different architectures to perform unsupervised representation learning using GAN are promising. Instead of using hand-engineered priors of sampling in the generator, the work in~\cite{nguyen2016plug} learned priors using denoising auto-encoders. For measuring the sample similarity for complex distributions such as those in the images, \cite{larsen2016autoencoding} jointly trained variational autoencoders and GANs. Moreover, the applications of CatGAN~\cite{springenberg2015unsupervised} and InfoGAN~\cite{chen2016infogan} for semi-supervised and unsupervised classification tasks in medical imaging are worth exploring as well.

Medical imaging has unique challenges associated with the scarcity of labeled examples. Moreover, unless corroborated by biopsy, there may exist a large variability in labeling from different radiologists. Although fine-tuning has helped to address the lack of annotated
examples, the performance is limited due to large differences in domains. It is comparatively easier to obtain scan level labels than slice level labels. In this regard, weakly supervised approaches such as multiple instance learning (MIL) can be of great value. Active learning can be another solution to alleviate the difficulty in labeling. In addition to these directions, unsupervised learning approaches will surely be pursued to address unique medical imaging challenges.

\bibliographystyle{splncs04}
\bibliography{tumor}
\end{document}